%% file: main.tex
\documentclass[runningheads]{llncs}
\usepackage[T1]{fontenc}
\usepackage{amsmath}
\usepackage{amsfonts}
\usepackage{amssymb}
\usepackage{wrapfig}
\usepackage{tikz}
\usepackage{comment}
\usepackage{amsmath,amssymb} 
\usepackage{color}
\usepackage{multirow}
\usepackage{kotex}
\usepackage{adjustbox}
\usepackage{float}
\usepackage{booktabs}
\usepackage{colortbl}
\usepackage{booktabs}

\def\Plus{\texttt{+}}
\newcommand{\myparagraph}[1]{\noindent{\bf #1}}

\usepackage{graphicx}
\newcommand*\samethanks[1][\value{footnote}]{\footnotemark[#1]}
\begin{document}
\title{Advancing Text-Driven Chest X-Ray Generation with Policy-Based Reinforcement Learning}
\titlerunning{Text-Driven Chest X-Ray Generation with Reinforcement Learning}
\author{Woojung Han\thanks{Equal contribution}, Chanyoung Kim\samethanks, Dayun Ju, \\ Yumin Shim, and Seong Jae Hwang\thanks{Corresponding author}}
\authorrunning{Han \textit{et al.}}

\institute{Yonsei University\\
\email{\tt\small \{dnwjddl, chanyoung, juda0707, symin0425, seongjae\}@yonsei.ac.kr}}

\maketitle              
\begin{abstract}

Recent advances in text-conditioned image generation diffusion models have begun paving the way for new opportunities in modern medical domain, in particular, generating Chest X-rays (CXRs) from diagnostic reports. 
Nonetheless, to further drive the diffusion models to generate CXRs that faithfully reflect the complexity and diversity of real data, it has become evident that a nontrivial learning approach is needed. 
In light of this, we propose CXRL, a framework motivated by the potential of reinforcement learning (RL). 
Specifically, we integrate a policy gradient RL approach with well-designed multiple distinctive CXR-domain specific reward models. 
This approach guides the diffusion denoising trajectory, achieving precise CXR posture and pathological details. 
Here, considering the complex medical image environment, we present ``RL with Comparative Feedback'' (RLCF) for the reward mechanism, a human-like comparative evaluation that is known to be more effective and reliable in complex scenarios compared to direct evaluation. 
Our CXRL framework includes jointly optimizing learnable adaptive condition embeddings (ACE) and the image generator, enabling the model to produce more accurate and higher perceptual CXR quality. 
Our extensive evaluation of the MIMIC-CXR-JPG dataset demonstrates the effectiveness of our RL-based tuning approach.
Consequently, our CXRL generates pathologically realistic CXRs, establishing a new standard for generating CXRs with high fidelity to real-world clinical scenarios.

\end{abstract}
\input{tex/introduction}
\input{tex/method}
\input{tex/experiment}
\input{tex/conclusion}

\newpage
\bibliographystyle{splncs04}
\bibliography{main}

\input{supple}

\end{document}

%% file: tex/introduction.tex
\section{Introduction}

\begin{figure*}[t!]
    \begin{center}
        \includegraphics[width=\textwidth]{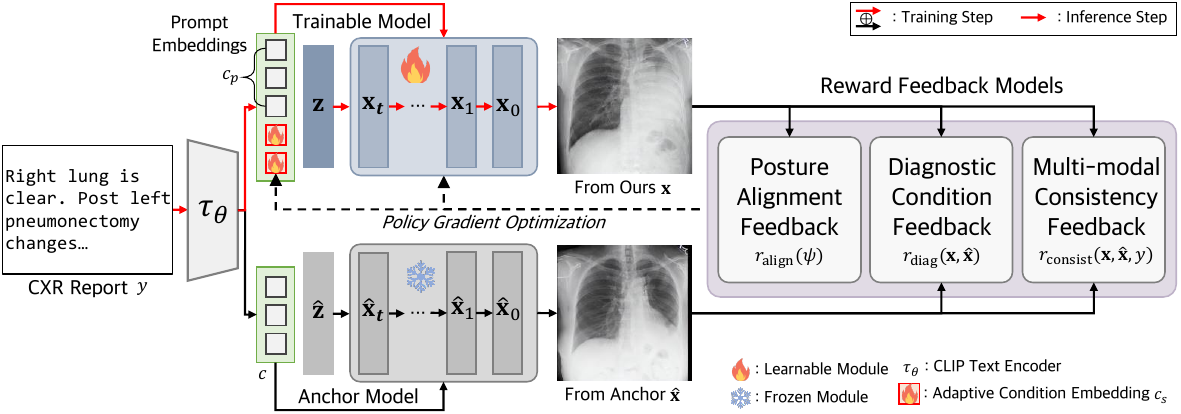}
    \end{center}
    \vspace{-17pt}
    \caption{The overview of \textit{\textbf{CXRL}}. Our model employs policy gradient optimization utilizing multi-reward feedbacks, fine-tuning image generator and ACE to produce realistic and accurate CXR that corresponds closely to the input report.
    }
    \vspace{-10pt}
    \label{fig:overview}
\end{figure*}

Diffusion probabilistic models (DPMs)~\cite{ho2020denoising,song2021denoising} have revolutionized conditional image synthesis~\cite{rombach2022high}, particularly in the realm of medical imaging~\cite{peng2022generating,du2023arsdm,pinaya2022brain,khader2022medical}, by enabling text-driven generation~\cite{chambon2022roentgen,jiang2023cola}.
This innovative approach addresses medical field challenges, such as data scarcity and privacy concerns, offering a promising avenue for synthetic medical data utilization in various applications (i.e., privacy-free medical image data for medical education or research)~\cite{KAZEROUNI2023102846}.

However, generating medical images via text conditions is challenging with diffusion models, mainly due to their inability to accurately represent subtle diagnostic differences, while medical images have complex characteristics (i.e., minor tissue texture variations per patient or disease)~\cite{KAZEROUNI2023102846}. 
This issue emerges because of the reliance on approximating the log-likelihood objective of the training data, prompting the model to predominantly generate images that align with the distribution of the training data~\cite{fan2023reinforcement,black2024training}, thus failing to accurately represent minor variations characteristic of medical images.
Meanwhile, recent computer vision studies suggest treating diffusion denoising as a multi-step decision-making problem~\cite{fan2023reinforcement,hao2023optimizing,black2024training,lee2024parrot}, allowing the use of policy gradient-based reinforcement learning (RL) which can overcome the existing limitations.
Motivated by such potential, we first integrate policy-based RL into medical image synthesis by tuning an image generator to enhance text-to-medical image generation using non-differentiable reward functions.
However, our empirical observations indicated that solely optimizing the image generator limits the scope of pathological expressions. 
Consequently, we introduce \textbf{CXRL}, jointly fine-tuning the image generator and \textit{learnable adaptive condition embeddings} (ACE) to condition the image generation process flexibly, enriching pathological features represented. 

Although prior works~\cite{fan2023reinforcement,hao2023optimizing,black2024training,lee2024parrot} have successfully incorporated data-driven human feedback models~\cite{ke2023vila,kirstain2023pickapic} into RL for image generation, applying human feedback models to medical image generation task presents unique and significant challenges.
These challenges primarily stem from the ``complexity of the medical imaging environment'', which makes it difficult for humans to quantitatively express the overall quality of medical images with absolute numbers~\cite{margaret2012practical,rosen2018quantitative}.
Given these challenges, particularly in the context of complex and uncertain environments like medical imaging, the findings of Mussweiler and Posten~\cite{MUSSWEILER2012236} stand out, illustrating that humans are better at making consistent and accurate judgments through ``relative comparison'' rather than absolute evaluation.

Therefore, we adopt \textit{Reinforcement Learning with ``Comparative'' Feedback} (RLCF) method, similar to~\cite{hao2023optimizing}, which enhances the reliability of the feedback mechanism by human-like comparative evaluation processes, a suitable feedback for complex text-to-medical image generation environments (Fig.~\ref{fig:overview}).
Specifically, RLCF assesses image pairs generated from ours and the anchor model, providing rewards for surpassing the anchor's images and penalties when our model's images fall short of the anchor's, which we categorize as negative actions.
This strategy employs both rewards and penalties, steering our model away from producing images inferior to the anchor's images. 
To sum up, our RLCF framework not only mimics human feedback for a reliable training experience but also integrates penalties to deter negative actions in synthesizing medical images.

Focusing specifically on Chest X-ray (CXR) generation, this paper addresses three key challenges in generating report-conditioned CXRs tackling each with specific feedback to achieve our final goal.
\textbf{(i)} Posture Alignment Feedback: We ensure accurate posture alignment through a specialized network, addressing potential misalignments that could impair disease detection and diagnostics, such as missing crucial areas (i.e., costophrenic angle). 
\textbf{(ii)} Diagnostic Condition Feedback: To enhance diagnostic accuracy in generated CXRs, we use feedback from a state-of-the-art pretrained CXR lesion classifier~\cite{cohen2022torchxrayvision} which guides the model to generate more precise pathologic features.
\textbf{(iii)} Multimodal Consistency Feedback: 
To ensure that CXRs generated from reports are semantically in agreement with the reports, we provide feedback using large-scale representations~\cite{you2023cxr} trained on CXR-report pairs. 
This feedback guides our CXR generation model to overcome disagreement between generated CXRs and reports.

\myparagraph{Contributions.}
\textbf{(1)} Our study pioneers in applying RL to text-conditioned medical image synthesis, particularly in CXRs, focusing on detail refinement and input condition control for clinical accuracy.
\textbf{(2)} We advance report-to-CXR generation with a RLCF-based rewarding mechanism, emphasizing posture alignment, pathology accuracy, and consistency between input reports and generated CXRs.
\textbf{(3)} We jointly optimize the image generator and ACE via reward feedback models, ensuring image-text alignment and medical accuracy across varied reports, setting a new benchmark in a report-to-CXR generation.

%% file: tex/method.tex
\section{Methods}
\begin{figure*}[t!]
    \begin{center}
        \includegraphics[width=\textwidth]{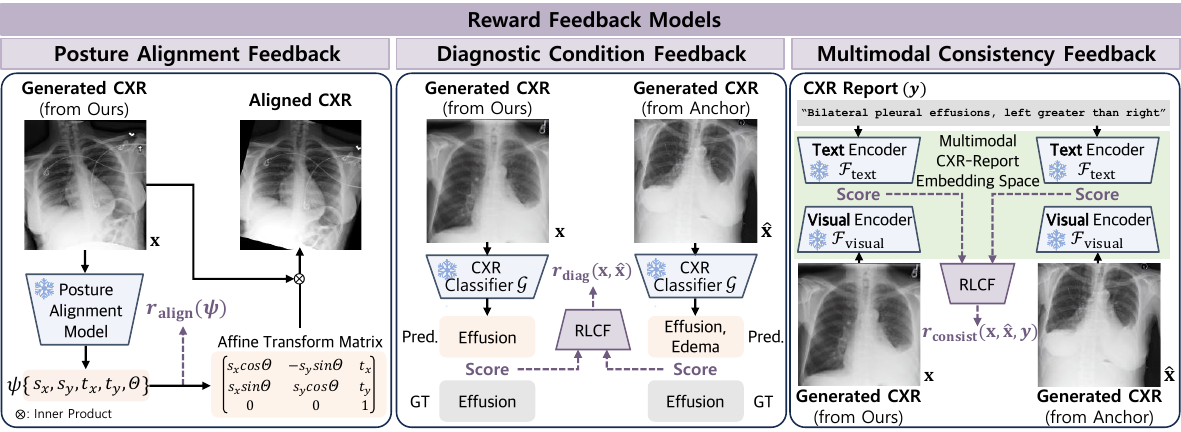} 
    \end{center}
    \vspace{-17pt}
    \caption{A detailed illustration of our reward feedback models. We incorporate three different feedbacks for report-to-CXR generation model to generate goal-oriented CXRs.
    }
    \vspace{-10pt}
    \label{fig:reward}
\end{figure*}

\vspace{-10pt}
In our study, we integrate policy gradient RL with diffusion models to enhance the denoising process.
We first describe the preliminaries and detail our framework (Fig.~\ref{fig:overview}) guided by a carefully designed multi-reward system (Fig.~\ref{fig:reward}).

\subsection{Preliminaries}

\myparagraph{Conditional Diffusion Probabilistic Models.}
Conditional Diffusion Probabilistic Models (Conditional DPMs)~\cite{rombach2022high} improve upon DPMs~\cite{ho2020denoising,song2021denoising} by integrating conditions like text prompts into the reverse diffusion process, guiding the synthesis towards high diversity and adherence to specified features. These models train on a conditional log-likelihood to adjust outputs closely to these conditions, enhancing diversity and targeted representation capability.

\myparagraph{RL-based Diffusion Model Fine-tuning.}
The RL paradigm interprets \textit{the diffusion denoising process as a series of actions} aiming to maximize a reward function that evaluates the adherence of generated images to a given prompt~\cite{black2024training}. 
The sequential decision-making (denoising process) is formulated by an objective function that targets the cumulative reward across the diffusion trajectory. 
Given the context of RL-based fine-tuning of conditional DPMs, the objective function $\mathcal{J}_\theta$ is defined as $ \mathcal{J}_\theta = \mathbb{E}_{p(\mathbf{y})}\mathbb{E}_{p_\theta(\mathbf{x}|\mathbf{y})}\left[r(\mathbf{x}, \mathbf{y})\right]$,
where $p(\mathbf{y})$ is the distribution of the text condition $\mathbf{y}$, $p_\theta(\mathbf{x}|\mathbf{y})$ is the distribution produced by the pretrained DPM conditioned on text $\mathbf{y}$, and $r(\mathbf{x}, \mathbf{y})$ is the reward for image $\mathbf{x}$ given text condition $\mathbf{y}$.
As in DPOK~\cite{fan2023reinforcement}, the policy gradient of $\mathcal{J}_\theta$ w.r.t. the parameters $\theta$ is 

\begin{equation}
\small
\label{policy_gradient_eq}
    \nabla_{\theta}\mathcal{J}_\theta = \mathbb{E}\left[r(\mathbf{x}, \mathbf{y}) \sum\nolimits_{t=1}^{T} \nabla_{\theta} \log p_\theta(\mathbf{x}_{t-1}|\mathbf{x}_t,\mathbf{y}, t)\right],
\end{equation}
where $T$ is the total number of time steps in the diffusion trajectory, $\mathbf{x}_t$ is the state of the image at time step $t$, and the gradient $\nabla_{\theta}$ is taken w.r.t. the model parameters $\theta$.

By applying policy gradients, DPMs iteratively fine-tune their parameters, improving image quality and enabling the generation of images with high fidelity, detail, and diversity.

\subsection{Reinforcement Learning with Comparison Feedback}
Unlike existing methods that rely on direct feedback~\cite{fan2023reinforcement,black2024training,lee2024parrot}, we employ Reinforcement Learning with Comparative Feedback (RLCF), contrasting model performance with a static anchor model, as illustrated in Fig.~\ref{fig:overview} and Fig.~\ref{fig:reward}. 
Utilizing RLCF, our model adapts through positive feedback for outperforming, and negative for underperformance, compared to the frozen anchor.
With given reward feedback signals, we jointly optimize two learning aspects:

\textbf{(I) Image Generator:} The image generator (diffusion model) aims to create the final CXR $\mathbf{x}$ through a series of denoising steps, conditioned by the text prompt embeddings $c$ extracted from CXR report $\mathbf{y}$. 
We fine-tune the diffusion layers in a RL manner, advancing their fidelity in medical imaging.

\textbf{(II) Learnable Adaptive Condition Embeddings: }
Given the variability in radiologist reporting styles, relying solely on a report itself for CXR generation presents significant challenges.
For this reason, inspired by Lester~\textit{et al.}~\cite{lester2021power}, we combine original static prompt embeddings $c_p = \tau_\theta(\mathbf{y})\in\mathbb{R}^{M\times d_\tau}$ with the \textit{learnable adaptive condition embeddings} (ACE), represented as $c_s\in\mathbb{R}^{N\times d_\tau}$. 
It is achieved by concatenating them to form the input condition as $c = [c_s, c_p]\in\mathbb{R}^{(N+M)\times d_\tau}$.
Here, $\tau_\theta$ symbolizes the parameterized frozen text encoder for conditioning image generator, $\textbf{y}$ is the input report, $d_\tau$ is the embedding dimension, $N$ is a sequence length of $c_s$ and $M$ denotes sequence length of $c_p$.
ACE and $c_p$ jointly allow the image generator more flexible inputs via cross-attention, enabling a precise capture of diagnostic and posture alignment features in CXRs.

\subsection{Goal-oriented Reward Feedback Models}
Our reward system comprises three models with distinct objectives (Fig.~\ref{fig:reward}).
\vspace{-10pt}

\subsubsection{(I) Posture Alignment Feedback.}
Generated CXRs often face scaling issues, like excessive zooming or rotation, obscuring essential details.
To counter these undesirable effects, we introduce a reward signal to align the CXR's posture with a canonical orientation to preserve essential parts.
The posture alignment model is trained using L2 loss between the real CXRs and canonical CXR, where canonical CXR is created by averaging 500 random CXRs from the dataset, following Liu \textit{et al.}~\cite{Liu_2019_ICCV}.
The posture alignment model adjusts CXRs by predicting parameters $\psi = \{s_x, s_y, t_x, t_y, \Theta\}$ for an affine transformation, including scaling $(s_x, s_y)$, translation $(t_x, t_y)$, and rotation $\Theta$. 
The posture alignment reward, $r_\text{align}(\psi)$, quantifies the magnitude of transformation obtained from the frozen posture alignment model.
Our $r_\text{align}$ is calculated in independent parameter space as $r_\text{align}(\psi) = -\{(\text{max}(|s_x-1|, |s_y-1|) + \frac{|\Theta|}{2\pi} + \sqrt{t_x^2 + t_y^2})\}$.
It aligns CXRs to the canonical pose, enhancing posture precision and visibility of essential regions.

\subsubsection{(II) Diagnostic Condition Feedback.}
To accurately reflect generated CXRs with referenced pathologies, we classify them using parsed report label, rewarding its accuracy. 
To keep away from negative actions, we integrate the RLCF, comparing our model's CXR $\mathbf{x}$ against anchor model's CXR $\mathbf{\hat{x}}$, using their score difference for reward. 
Our diagnostic condition feedback $r_\text{diag}$ is defined by the formula: $r_\text{diag}(\mathbf{x}, \mathbf{\hat{x}}) = \{\text{accuracy}(\mathcal{G}(\mathbf{x})) - \text{accuracy}(\mathcal{G}(\mathbf{\hat{x}}))\}$, where $\mathcal{G}$ represents a state-of-the-art pretrained CXR lesion classifier network~\cite{cohen2022torchxrayvision}.

\subsubsection{(III) Multimodal Consistency Feedback.}
Finally, we enforce the generated CXRs to better match their reports.
We leverage a multimodal latent representation~\cite{you2023cxr} pretrained with CXR-report pairs for semantic agreement assessment. 
Specifically, we employ RLCF to compute the difference in similarity scores between generated CXR $\mathbf{x}$ by our model and the report $\textbf{y}$, and the generated CXR $\mathbf{\hat{x}}$ by an anchor model and the same report. 
The reward feedback is defined as $r_\text{consist}(\mathbf{x}, \mathbf{\hat{x}}, \mathbf{y}) = \{d_\text{cosine}(\mathcal{F}_\text{visual}(\mathbf{x}), \mathcal{F}_\text{text}(\mathbf{y})) - d_\text{cosine}(\mathcal{F}_\text{visual}(\mathbf{\hat{x}}), \mathcal{F}_\text{text}(\mathbf{y})) \}$, where $d_\text{cosine}$ represents the cosine distance and $\mathcal{F}_\text{visual}$ and $\mathcal{F}_\text{text}$ represent the visual and text encoders, respectively, mapping feature embeddings from their respective modal inputs to joint multimodal latent space~\cite{you2023cxr}.

\begin{figure*}[t!]
    \begin{center}
        \includegraphics[width=\textwidth]{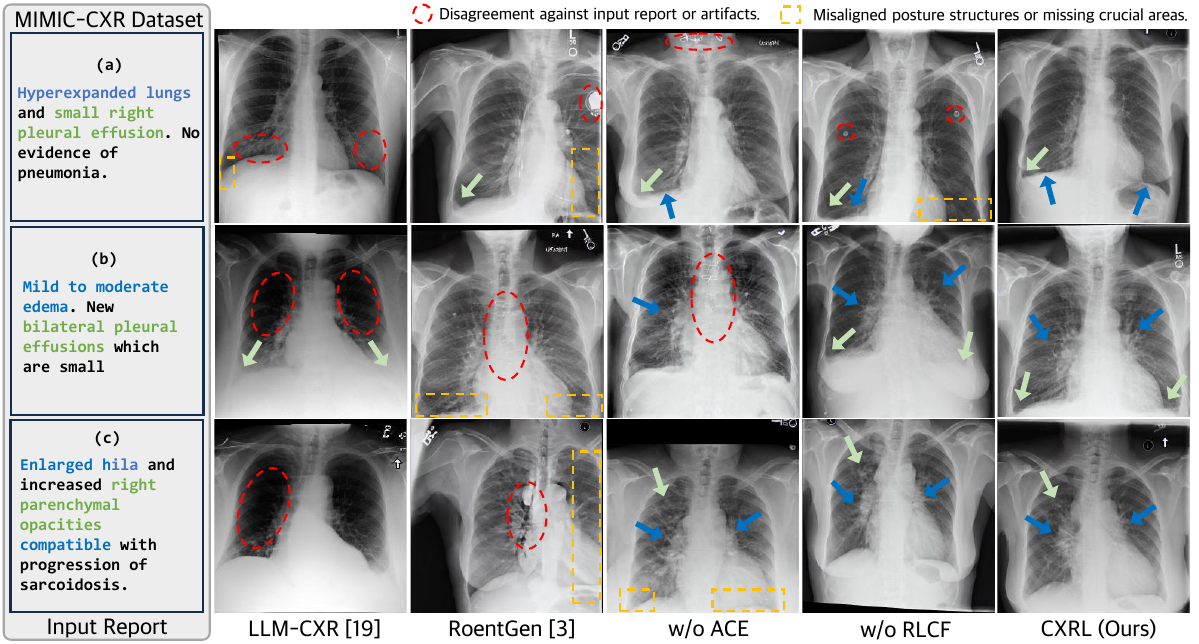} 
    \end{center}
    \vspace{-17pt}
    \caption{Comparison between previous state-of-the-art report-to-CXR generation models~\cite{lee2024llmcxr,chambon2022roentgen} and ours.  
    The blue and green texts match their corresponding colored arrows.}
    \label{fig:qualitative}
    \vspace{-10pt}
\end{figure*}

\subsection{Optimizing Policy Gradient} 
Integrating the introduced reward feedbacks, we update the policy gradient in a RL manner, by expanding Eq.~\eqref{policy_gradient_eq}.
Our final reward to maximize is $r(\mathbf{x}, \mathbf{\hat{x}}, \mathbf{y}, \psi) = \lambda_\text{align} r_\text{align}(\psi) + \lambda_\text{diag} r_\text{diag}(\mathbf{x}, \mathbf{\hat{x}}) + \lambda_\text{consist}r_\text{consist}(\mathbf{x}, \mathbf{\hat{x}}, \mathbf{y})$, where $\lambda_\text{align}$, $\lambda_\text{diag}$, and $\lambda_\text{consist}$ are hyperparameters. With given rewards, for $B$ images in mini-batches and $T$ total diffusion denoising steps, we update the gradient as follows:
\begin{equation}
\small
    \begin{aligned}
        \nabla \mathcal{J}_\theta = {1 \over B}\sum_{i=1}^{B}\sum^T_{t=1}(r(\mathbf{x}^{(i)}, \mathbf{\hat{x}}^{(i)}, \mathbf{y}^{(i)}, \psi^{(i)})) \times \nabla_\theta \log p_\theta({\bf x}_{t-1}^{(i)} | {\bf x}_t^{(i)}, c^{(i)}, t).
    \end{aligned}
    \label{eq:optimize}
\end{equation}

%% file: tex/experiment.tex
\section{Experiments}

\subsubsection{Datasets.}
We utilize PA and AP views from the MIMIC-CXR-JPG dataset~\cite{johnson2019mimic_jpg}, while limiting report impressions to 74 tokens or less. 
We randomly sample 5,000 pairs from the official train split for training.
For testing, similarly to the training set, we utilize PA views from the test split, resulting in 1,000 pairs.

\myparagraph{Implementation Details.}
CXRL fine-tunes with LoRA layers~\cite{hu2022lora} on RoentGen-initialized Stable Diffusion~\cite{chambon2022roentgen,rombach2022high}, whereas the anchor model uses static RoentGen.
We train a posture alignment model with the MIMIC-CXR-JPG dataset~\cite{johnson2019mimic_jpg} using ResNet-18~\cite{he2016deep}.
A DenseNet-121~\cite{huang2017densely} CXR classifier~\cite{cohen2022torchxrayvision} provides diagnostic feedback.
$\mathcal{F}_\text{text}$ and $\mathcal{F}_\text{vision}$ utilize CXR-CLIP~\cite{you2023cxr}, with BioClinicalBERT~\cite{alsentzer2019publicly} and ResNet50~\cite{he2016deep} serving as the backbone, respectively.
DDPO algorithm~\cite{black2024training} guides policy gradient optimization, with 81 CXRs per batch, learning rate of 3e-4. 
$\lambda_\text{align}$, $\lambda_\text{diag}$, and $\lambda_\text{consist}$ are at 1, 10, and 10, with $N=3$ and $M\leq74$. 

\subsection{Evaluation Results}

\myparagraph{Qualitative Analysis.}
In Fig.~\ref{fig:qualitative}, we qualitatively compare our method to previous models~\cite{chambon2022roentgen,lee2024llmcxr}. 
\textbf{(1) \textit{Disagreement against input report or artifacts (red circle):}}
Our CXRL exhibits superior agreement between CXR and report compared to the baselines, especially in their capability to avoid generating unmentioned objects. 
For instance, in Fig.~\ref{fig:qualitative}(a), while RoentGen deviates from the report content by generating a pacemaker, CXRL maintains semantic consistency by not doing so.
Furthermore, in case Fig.~\ref{fig:qualitative}(b), RoentGen generates sternal wires (red circle) that are not mentioned in the report, whereas CXRL does not. 
In terms of diagnosing specific conditions, CXRL shows a high level of agreement. 
For example, in case Fig.~\ref{fig:qualitative}(c), while LLM-CXR and RoentGen either fail to identify or incorrectly generate hilar enlargement (blue colored text) with artifacts, CXRL accurately represents them (blue arrow).
\textbf{(2) \textit{Misaligned posture structures or missing crucial areas (yellow box):}}
While others occasionally miss details, our CXRL consistently shows the thoracic anatomy comprehensively, including lung apices and costophrenic angles.
In contrast, RoentGen often loses critical areas due to inadequate posture alignment.

\begingroup
\setlength{\tabcolsep}{6pt}
\renewcommand{\arraystretch}{0.9} 
\begin{table}[t!]
\caption{Evaluation of generated CXRs from multiple feedback perspectives. The table compares the performance of various methods using three evaluation metrics.}
    \centering\resizebox{\textwidth}{!}{
    \small
    \begin{tabular}{l c c c c}
        \toprule
        Method & (a) Posture Alignment & \multicolumn{1}{c}{(b) Classification} & \multicolumn{1}{c}{(c) Semantic Consistency} \\
        &  $r_\text{align}(\psi)$ $(\uparrow)$ & AUROC $(\uparrow)$ & CXR-CLIP Distance $(\uparrow)$ \\
        \midrule
        RoentGen~\cite{chambon2022roentgen} & -0.363 & 0.65 & 0.285 \\
        LLM-CXR~\cite{lee2024llmcxr} & -0.327 & \underline{0.78} & 0.201\\
        \midrule 
        \textbf{CXRL \textit{(Ours)}}  & \textbf{-0.246} & \textbf{0.81}  & \textbf{0.343}\\
        \midrule 
        \textit{w/o} ACE  & -0.354 & 0.66 & \underline{0.321} \\
       \textit{w/o} RLCF (direct feedback) & \underline{-0.319} & 0.71 & 0.301 \\
        \bottomrule
    \end{tabular}}
    \label{Tab.quantitative1}
    \vspace{-10pt}
\end{table}
\endgroup

\myparagraph{Quantitative Evaluation on CXR Specific Quality.}
Table~\ref{Tab.quantitative1} shows the evaluation of our generated CXR across three metrics, reflecting the rewards we targeted, which are key qualities of CXR. 
(a) The $r_\text{align}$ measures posture alignment, where values closer to 0 indicate a better fit. 
(b) The AUROC score assesses diagnostic accuracy, and (c) the semantic consistency score evaluates the alignment between CXR and the report. 
Table~\ref{Tab.quantitative1} confirms our rewards are enhancing report-to-CXR generation model towards the goal-oriented result.

\myparagraph{Quantitative Evaluation on Image Quality.}
Table~\ref{Tab.quantitative2}a details the fidelity of our generated CXRs, using the FID and MS-SSIM metrics for benchmarking against baselines. 
Employing a CXR-specific DenseNet-121~\cite{cohen2022torchxrayvision} for FID, CXRL outperforms baseline models~\cite{chambon2022roentgen,lee2024llmcxr}, showcasing superior image realism.
Our MS-SSIM scores, indicative of structural accuracy, also prove competitive, affirming our model's performance in generating authentic CXR textures.

\begin{table}[!t]

    \caption{Comparative analysis of generated CXR quality: (a) quantitatively compares established models using FID and MS-SSIM metrics; (b) evaluates the impact of reward components on FID scores.     }
    \begin{minipage}[b]{0.50\textwidth}
        \centering
        \renewcommand{\arraystretch}{0.15} 
        \setlength{\tabcolsep}{6pt} 
        \begin{tabular}{lcc}
            \multicolumn{3}{c}{(a) Quality Assessment} \\
            \toprule
            Model & FID $(\downarrow)$ & MS-SSIM $(\downarrow)$ \\
            \midrule
            RoentGen~\cite{chambon2022roentgen} & 2.58 & 0.197 \\
            LLM-CXR~\cite{lee2024llmcxr} & 4.01 & 0.275 \\
            \midrule
            \textbf{CXRL \textit{(Ours)}} & \textbf{1.51} & \textbf{0.147} \\
            \midrule
            \textit{w/o} ACE & 1.79 & 0.192 \\
            \textit{w/o} RLCF & \underline{1.58} & \underline{0.188} \\
            \bottomrule
        \end{tabular}
        \label{Tab.quantitative2}
    \end{minipage}
    \hfill
    \begin{minipage}[b]{0.45\textwidth}
        \centering
        \renewcommand{\arraystretch}{0.9} 
        \setlength{\tabcolsep}{0pt}
        \begin{tabular}{c|lc}
            \multicolumn{3}{c}{(b) Ablation: Reward Feedback Models} \\
            \toprule
            Exp.\# & \ Reward & FID $(\downarrow)$ \\
            \midrule
            - & \ Anchor Model~\cite{chambon2022roentgen} &  2.58 \\
            1 & \ $+ \ r_\text{align}$ &  1.89 \\
            2 & \ $+ \ r_\text{diag}$ & 1.64 \\
            3 & \ $+ \ r_\text{consist}$ & 1.92 \\
            \midrule
            \textbf{4} & \ \textbf{$r_\text{align} + r_\text{diag} + r_\text{consist}$} & \textbf{1.51} \\
            \bottomrule
        \end{tabular}
        \label{Tab.ablation}
    \end{minipage}
    \vspace{-10pt}
\end{table}

\myparagraph{Medical Expert Assessment.}
In pursuit of medical expert feedback, we assessed 20 generated CXRs with five radiology experts. 
They rated generated CXRs from RoentGen, LLM-CXR, and CXRL against reports on a 1 to 5 scale. 
Average scores were:  RoentGen \textbf{3.4}, LLM-CXR \textbf{2.4}, and CXRL at \textbf{4.1}, showing superior CXR-report alignment and realism of our generated CXRs.

\subsection{Ablation Study}

\myparagraph{Effect of ACE and RLCF.}
We validate the effectiveness of ACE (Adaptive Condition Embeddings) and RLCF (comparative reward feedback) by excluding the corresponding factor from fully combined CXRL.
Table~\ref{Tab.quantitative1} presents a comparison of key CXR quality metrics in our CXRL framework with and without the integration of ACE and RLCF.
While removing either ACE or RLCF improves performance over the anchor model~\cite{chambon2022roentgen}, ours with both ACE and RLCF achieves the best overall performance. 
Similar aspects are shown in terms of image quality assessment (Table~\ref{Tab.quantitative2}(a)).
In particular, w/o RLCF, which uses direct feedback optimization with ACE, seems to improve CXR quality compared to the anchor by providing flexible input conditions. 
This experiment demonstrates that both ACE and RLCF are essential contributions of our model, in both domain-specific quality and image quality aspects (See Fig.~\ref{fig:qualitative} for qualitative result).

\myparagraph{Effect of Reward Models.}
In Table~\ref{Tab.ablation}(b), we detail how each reward signal influences model performance by examining FID scores. 
Individual rewards enhance CXR quality (Exp. \#1 to \#3) beyond anchor levels which we aim to surpass. 
Combining all rewards (Exp. \#4) synergistically lowers the FID, showing that multiple feedback signals crucially enhance CXR synthesis by capturing broader features for high-fidelity CXRs
(See supplement for qualitative result).

%% file: tex/conclusion.tex
\section{Conclusion and Discussion}
In this study, we present \textbf{CXRL} for report-to-CXR generation using RLCF, which significantly enhances medical image synthesis through a policy gradient RL and an advanced adaptable reward feedback system. 
Our exhaustive experiments demonstrate our framework's state-of-the-art performance in generating CXRs. 
The proposed adaptable reward framework not only excels in CXR generation but may also pioneer a wide spectrum of applications in various domains of medical image synthesis, offering enhanced diagnostics, contributing to medical education and research, and ensuring ethical access to medical imaging data.

%% file: supple.tex

\def\Plus{\texttt{+}}

\title{Supplementary Materials, Advancing Text-Driven Chest X-Ray Generation With Policy-Based Reinforcement Learning}

\author{Woojung Han\thanks{Equal contribution}, Chanyoung Kim\samethanks, Dayun Ju, \\ Yumin Shim, and Seong Jae Hwang\thanks{Corresponding author}}
\authorrunning{Han \textit{et al.}}

\institute{Yonsei University}
\maketitle              

\begin{figure*}[h]
    \begin{center}
        \includegraphics[width=\textwidth]{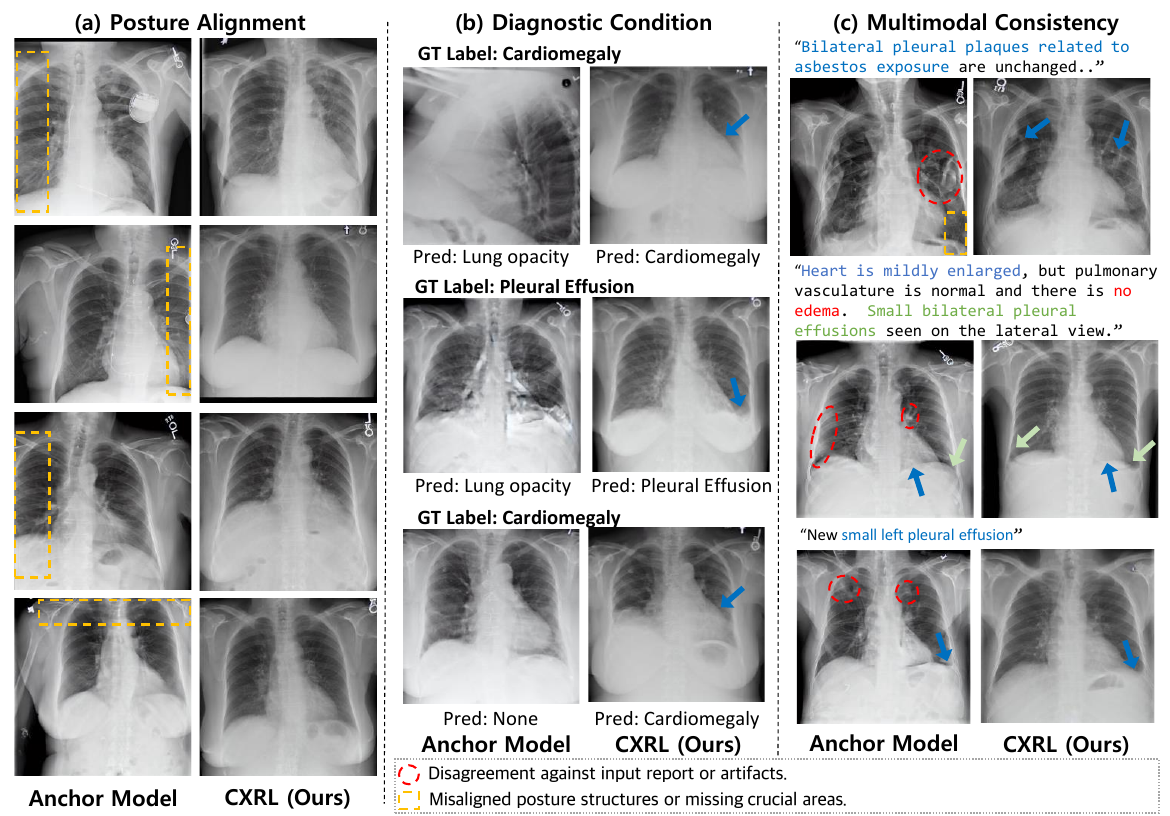}
    \end{center}
    \vspace{-10pt}
    \caption{Qualitative ablation on each reward models. (a): CXRL shows significantly better alignment of the \textit{clavicle} and \textit{costophrenic angle} compared to anchor regarding posture alignment. (b): CXRL demonstrates improved predictive diagnostic accuracy, closely matching the GT and enhancing clinical decision-making. (c): The multimodal consistency reward ensures that CXRs and reports correspond well, as observed by arrows and text in matching colors. Conversely, anchor sometimes generates content in the report that is either irrelevant or should not be present. Specifically, in the second row, the presence of edema is verifiable within the red circle, indicating misalignment between the CXR and the report.
    }
\end{figure*}

\begin{table}[h]
\caption{AUROC scores for each pathology class, obtained by processing generated CXRs from various models through a state-of-the-art CXR pathology classifier. 
Notably, our \textbf{CXRL} model outperforms other baselines and even surpasses the diagnostic accuracy of original real CXRs (GT) in the dataset, highlighting the proficiency of our approach in capturing the details of each pathology accurately.
}
\vspace{-10pt}
\begin{center}
\addtolength{\tabcolsep}{-1.1pt}  
\small
\begin{tabular}{cccccccccccc|c}
\toprule
  \textbf{AUROC} ($\uparrow$) & \textbf{Atel.} & \textbf{Cnsl.} & \textbf{Pmtx.} & \textbf{Edem.}  & \textbf{Eff.} & \textbf{Pna.} & \textbf{Cmgl.} & \textbf{Les.} & \textbf{Frac.} & \textbf{Opac.} & \textbf{ECm.} & \textbf{Avg.}  \\
\midrule
RoentGen & .78 & .90 & .79 & .71 & .46 & .39 & .46 & .57 & .90 & .48 & .71 & .65 \\
LLM-CXR & .86  & .88 & .81  & .86 & .68  & \textbf{.76} & \textbf{.67}  & .76  & .88  & .62 & .82 &  .78  \\
\midrule
\textbf{CXRL} & \textbf{.87} & \textbf{.94} & \textbf{.88} & \textbf{.89} & \textbf{.77} & .60 & .63 & \textbf{.79} & \textbf{.94} & \textbf{.70} & \textbf{.86} & \textbf{.81} \\
\bottomrule
\vspace{-10pt}
\end{tabular}
\end{center}
\end{table}

\begin{figure*}[h]
    \begin{center}
        \includegraphics[width=\textwidth]{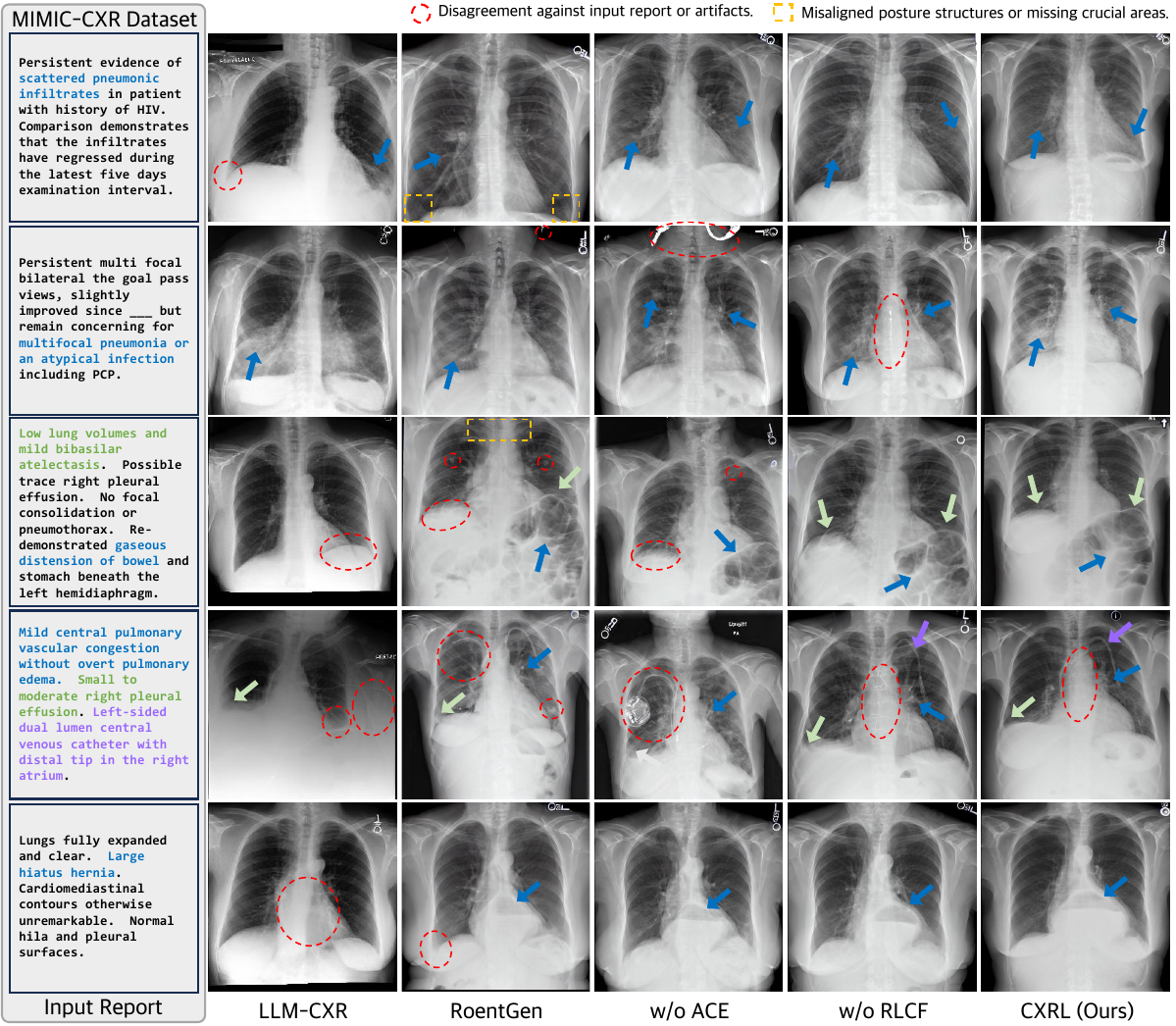} 
    \end{center}
    \vspace{-10pt}
    \caption{Additional qualitative results of our framework comparing against baselines. The colored texts match their corresponding colored arrows. 
    Ours w/o ACE or RLCF demonstrates superior report agreement and posture alignment compared to other baselines, and \textbf{CXRL} is observed to generate more advanced high-fidelity CXRs that highlight our methodology's effectiveness in synthesizing clinically accurate medical images.
    }
\end{figure*}

%% file: main.bbl
\begin{thebibliography}{10}
\providecommand{\url}[1]{\texttt{#1}}
\providecommand{\urlprefix}{URL }
\providecommand{\doi}[1]{https://doi.org/#1}

\bibitem{alsentzer2019publicly}
Alsentzer, E., Murphy, J., Boag, W., Weng, W.H., Jin, D., Naumann, T., McDermott, M.: Publicly available clinical {BERT} embeddings. In: Proceedings of the 2nd Clinical Natural Language Processing Workshop. pp. 72--78. Association for Computational Linguistics, Minneapolis, Minnesota, USA (Jun 2019)

\bibitem{black2024training}
Black, K., Janner, M., Du, Y., Kostrikov, I., Levine, S.: Training diffusion models with reinforcement learning. In: The Twelfth International Conference on Learning Representations (2024)

\bibitem{chambon2022roentgen}
Chambon, P., Bluethgen, C., Delbrouck, J.B., Van~der Sluijs, R., Po{\l}acin, M., Chaves, J.M.Z., Abraham, T.M., Purohit, S., Langlotz, C.P., Chaudhari, A.: Roentgen: vision-language foundation model for chest x-ray generation. arXiv preprint arXiv:2211.12737  (2022)

\bibitem{cohen2022torchxrayvision}
Cohen, J.P., Viviano, J.D., Bertin, P., Morrison, P., Torabian, P., Guarrera, M., Lungren, M.P., Chaudhari, A., Brooks, R., Hashir, M., et~al.: Torchxrayvision: A library of chest x-ray datasets and models. In: International Conference on Medical Imaging with Deep Learning. pp. 231--249. PMLR (2022)

\bibitem{du2023arsdm}
Du, Y., Jiang, Y., Tan, S., Wu, X., Dou, Q., Li, Z., Li, G., Wan, X.: Arsdm: colonoscopy images synthesis with adaptive refinement semantic diffusion models. In: International conference on medical image computing and computer-assisted intervention. pp. 339--349. Springer (2023)

\bibitem{fan2023reinforcement}
Fan, Y., Watkins, O., Du, Y., Liu, H., Ryu, M., Boutilier, C., Abbeel, P., Ghavamzadeh, M., Lee, K., Lee, K.: Reinforcement learning for fine-tuning text-to-image diffusion models. In: Thirty-seventh Conference on Neural Information Processing Systems (2023)

\bibitem{hao2023optimizing}
Hao, Y., Chi, Z., Dong, L., Wei, F.: Optimizing prompts for text-to-image generation. In: Thirty-seventh Conference on Neural Information Processing Systems (2023)

\bibitem{he2016deep}
He, K., Zhang, X., Ren, S., Sun, J.: Deep residual learning for image recognition. In: Proceedings of the IEEE conference on computer vision and pattern recognition. pp. 770--778 (2016)

\bibitem{ho2020denoising}
Ho, J., Jain, A., Abbeel, P.: Denoising diffusion probabilistic models. Advances in neural information processing systems  \textbf{33},  6840--6851 (2020)

\bibitem{hu2022lora}
Hu, E.J., yelong shen, Wallis, P., Allen-Zhu, Z., Li, Y., Wang, S., Wang, L., Chen, W.: Lo{RA}: Low-rank adaptation of large language models. In: International Conference on Learning Representations (2022)

\bibitem{huang2017densely}
Huang, G., Liu, Z., Van Der~Maaten, L., Weinberger, K.Q.: Densely connected convolutional networks. In: Proceedings of the IEEE conference on computer vision and pattern recognition. pp. 4700--4708 (2017)

\bibitem{jiang2023cola}
Jiang, L., Mao, Y., Wang, X., Chen, X., Li, C.: Cola-diff: Conditional latent diffusion model for multi-modal mri synthesis. In: International Conference on Medical Image Computing and Computer-Assisted Intervention. pp. 398--408. Springer (2023)

\bibitem{johnson2019mimic_jpg}
Johnson, A.E., Pollard, T.J., Greenbaum, N.R., Lungren, M.P., Deng, C.y., Peng, Y., Lu, Z., Mark, R.G., Berkowitz, S.J., Horng, S.: Mimic-cxr-jpg, a large publicly available database of labeled chest radiographs. arXiv preprint arXiv:1901.07042  (2019)

\bibitem{KAZEROUNI2023102846}
Kazerouni, A., Aghdam, E.K., Heidari, M., Azad, R., Fayyaz, M., Ilker: Diffusion models in medical imaging: A comprehensive survey. Medical Image Analysis  \textbf{88},  102846 (2023)

\bibitem{ke2023vila}
Ke, J., Ye, K., Yu, J., Wu, Y., Milanfar, P., Yang, F.: Vila: Learning image aesthetics from user comments with vision-language pretraining. In: Proceedings of the IEEE/CVF Conference on Computer Vision and Pattern Recognition. pp. 10041--10051 (2023)

\bibitem{khader2022medical}
Khader, F., Mueller-Franzes, G., Arasteh, S.T., Han, T., Haarburger, C., Schulze-Hagen, M., Schad, P., Engelhardt, S., Baessler, B., Foersch, S., et~al.: Medical diffusion--denoising diffusion probabilistic models for 3d medical image generation. arXiv preprint arXiv:2211.03364  (2022)

\bibitem{kirstain2023pickapic}
Kirstain, Y., Polyak, A., Singer, U., Matiana, S., Penna, J., Levy, O.: Pick-a-pic: An open dataset of user preferences for text-to-image generation. In: Thirty-seventh Conference on Neural Information Processing Systems (2023)

\bibitem{lee2024parrot}
Lee, S.H., Li, Y., Ke, J., Yoo, I., Zhang, H., Yu, J., Wang, Q., Deng, F., Entis, G., He, J., et~al.: Parrot: Pareto-optimal multi-reward reinforcement learning framework for text-to-image generation. arXiv preprint arXiv:2401.05675  (2024)

\bibitem{lee2024llmcxr}
Lee, S., Kim, W.J., Chang, J., Ye, J.C.: {LLM}-{CXR}: Instruction-finetuned {LLM} for {CXR} image understanding and generation. In: The Twelfth International Conference on Learning Representations (2024)

\bibitem{lester2021power}
Lester, B., Al-Rfou, R., Constant, N.: The power of scale for parameter-efficient prompt tuning. In: Proceedings of the 2021 Conference on Empirical Methods in Natural Language Processing. pp. 3045--3059. Association for Computational Linguistics, Online and Punta Cana, Dominican Republic (Nov 2021)

\bibitem{Liu_2019_ICCV}
Liu, J., Zhao, G., Fei, Y., Zhang, M., Wang, Y., Yu, Y.: Align, attend and locate: Chest x-ray diagnosis via contrast induced attention network with limited supervision. In: Proceedings of the IEEE/CVF International Conference on Computer Vision (ICCV) (October 2019)

\bibitem{margaret2012practical}
Margaret~Cheng, H.L., Stikov, N., Ghugre, N.R., Wright, G.A.: Practical medical applications of quantitative mr relaxometry. Journal of Magnetic Resonance Imaging  \textbf{36}(4),  805--824 (2012)

\bibitem{MUSSWEILER2012236}
Mussweiler, T., Posten, A.C.: Relatively certain! comparative thinking reduces uncertainty. Cognition  \textbf{122}(2),  236--240 (2012)

\bibitem{peng2022generating}
Peng, W., Adeli, E., Zhao, Q., Pohl, K.M.: Generating realistic 3d brain mris using a conditional diffusion probabilistic model. In: International conference on medical image computing and computer-assisted intervention. Springer (2023)

\bibitem{pinaya2022brain}
Pinaya, W.H., Tudosiu, P.D., Dafflon, J., Da~Costa, P.F., Fernandez, V., Nachev, P., Ourselin, S., Cardoso, M.J.: Brain imaging generation with latent diffusion models. In: MICCAI Workshop on Deep Generative Models. pp. 117--126. Springer (2022)

\bibitem{rombach2022high}
Rombach, R., Blattmann, A., Lorenz, D., Esser, P., Ommer, B.: High-resolution image synthesis with latent diffusion models. In: Proceedings of the IEEE/CVF conference on computer vision and pattern recognition. pp. 10684--10695 (2022)

\bibitem{rosen2018quantitative}
Rosen, A.F., Roalf, D.R., Ruparel, K., Blake, J., Seelaus, K., Villa, L.P., Ciric, R., Cook, P.A., Davatzikos, C., Elliott, M.A., et~al.: Quantitative assessment of structural image quality. Neuroimage  \textbf{169},  407--418 (2018)

\bibitem{song2021denoising}
Song, J., Meng, C., Ermon, S.: Denoising diffusion implicit models. In: International Conference on Learning Representations (2021)

\bibitem{you2023cxr}
You, K., Gu, J., Ham, J., Park, B., Kim, J., Hong, E.K., Baek, W., Roh, B.: Cxr-clip: Toward large scale chest x-ray language-image pre-training. In: International Conference on Medical Image Computing and Computer-Assisted Intervention. pp. 101--111. Springer (2023)

\end{thebibliography}
